# Sampling - Variational Auto Encoder - Ensemble: In the Quest of Explainable Artificial Intelligence

Sarit Maitra
*Alliance Business School*
*Alliance University*
Bengaluru, India
sarit.maitra@gmail.com

Vivek Mishra
School of Applied mathematics
*Alliance University*
Bengaluru, India
vivek.mishra@alliance.edu.in

Pratima Verma
Alliance Business School
*Alliance University*
Bengaluru, India
pratima.verma@alliance.edu.in

Manav Chopra
*Alliance Business School*
*Alliance University*
Bengaluru, India
cmanavbba721@bus.alliance.edu.in

Priyanka Nath
*Alliance Business School*
*Alliance University*
Alliance Business School
npriyankamba722@bus.alliance.edu.in

*Abstract—* Explainable Artificial Intelligence (XAI) models have recently attracted a great deal of interest from a variety of application sectors. Despite significant developments in this area, there are still no standardized methods or approaches for understanding AI model outputs. A systematic and cohesive framework is also increasingly necessary to incorporate new techniques like discriminative and generative models to close the gap. This paper contributes to the discourse on XAI by presenting an empirical evaluation based on a novel framework: Sampling - Variational Auto Encoder (VAE) - Ensemble Anomaly Detection (SVEAD). It is a hybrid architecture where VAE combined with ensemble stacking and SHapley Additive exPlanations is used for imbalanced classification. The finding reveals that combining ensemble stacking, VAE, and SHAP can not only lead to better model performance but also provide an easily explainable framework. This work has used SHAP combined with Permutation Importance and Individual Conditional Expectations to create a powerful interpretability of the model. The finding has an important implication in the real world, where the need for XAI is paramount to boost confidence in AI applications.

*Keywords—discriminative model; explainable artificial intelligence; ensemble stacking; generative model; shapley additive explanations;*

TABLE 1. ABREVIATION

| Term | Abbreviation |
|---|---|
| Area Under the Precision-Recall Curve | AUPRC |
| Artificial Intelligence | AI |
| Bernoulli Distribution | BD |
| Brier Score | BS |
| Cohen's Kappa Coefficient | CKC |
| Cross Validation | CV |
| Decision Tree | RF |
| Deep Learning | DL |
| Discriminative Models | DML |
| Ensemble Modelling | EM |
| Ensemble Stacking | ES |
| Ensemble Voting | EV |
| Explainable Artificial Intelligence | XAI |
| Evidence Lower Bound | ELBO |
| Gaussian Distribution | GD |
| Generative AI | GAI |
| Individual Conditional Expectation | ICE |
| K-nearest neighbor | KNN |
| Kullback-Leibler | KL |
| Logistic Regression | Log Reg |
| Machine Learning | ML |
| Matthew's Correlation Coefficient | MCC |
| Neural Network | NN |
| Permutation Importance Plot | PIP |
| Principal Component Analysis | PCA |
| Random Undersampler | RUS |
| Receiver Operating Characteristic - Area Under the Curve | ROCAUC |
| Sampling-VAE-Ensemble Anomaly Detection | SVEAD |
| SHapley Additive exPlanations | SHAP |
| Standard Deviation | SD |
| Support Vector Classifier | SVC |
| Synthetic Minority Oversampling Technique | SMOTE |
| Synthetic Minority Over-sampling Technique and Tomek Links | SMOTETomek |
| Truncated Singular Value Decomposition | t-SVD |
| t-distributed Stochastic Neighbor Embedding | t-SNE |
| Variational Auto Encoders | VAE |

## I. INTRODUCTION

The increasing complexity of ML models has led to a growing interest in XAI. While today's Industry 4.0 emphasizes smart and intelligent processes powered by technology, complex models, such as EM and DL approaches, have emerged as key technologies for accomplishing the goal ([23]; [33]). However, these models are often difficult to understand and trust, limiting their practical use in real-world applications ([2], [13]). To address this challenge, recent advances in ML have focused on constructing and leveraging internal representations within ML models [16]. These advances aim to enhance the interpretability and explainability of complex models, making them more accessible and trustworthy for deployment in real-world settings.

This study aims to address this issue by using DML and GAI on a skewed anomaly detection dataset. DML are concerned with classifying or predicting specific outcomes given input data, whereas GAI models are concerned with learning the data distribution itself to generate new data points. Both these AI models can benefit from XAI techniques, albeit in diverse ways. For DML, XAI helps in understanding and interpreting their predictions, while for GAI, XAI can assist in understanding the data generation process and identifying anomalies. GAI models like VAEs and DMLs can offer a degree of interpretability, but it is important to clarify the extent of their interpretability and how it differs from other models.

Bias often exists in class imbalances, and traditional model interpretation methods, though easy to understand, lack practicality in explaining them to businesses and determining success criteria. To be realistic, interpretation should provide knowledge of the model's operations, predictions, discrimination rules, or potential disruptions [16]. This study explores best practices for integrating hybrid models combining DML and GAI approaches in real-world applications, aiming to improve the interpretability and robustness of anomaly detection systems.

This study presents a novel SVEAD framework for anomaly detection, utilizing sampling, VAE for compressed data representation, and supervised algorithms for classification, offering an integrated approach considering unique data characteristics and advanced DL ML techniques. Table 1 lists the technical abbreviations used in this article. There is currently no tangible mathematical concept for interpretability or explainability, nor have they been analyzed by some metric. Both terminologies are used interchangeably in this article.

## II. PREVIOUS WORK

The hybridization of GAI and DML techniques has emerged as a promising avenue for enhancing the accuracy of AI systems. Numerous researchers have advocated for the advantages of integrating these two paradigms (e.g., [1]; [33]; [28]; [27]; [43]; and others). However, the adoption of these hybrid approaches within the business sector remains cautious, primarily due to challenges associated with interpretability.

While a substantial body of research has been dedicated to exploring the performance of various DML and GAI models on identical datasets (e.g., [4]; [36]; [38]; [14]; [3]; [31]; [18]), these efforts have often focused on quantitative assessments. Although some researchers have undertaken comparative evaluations of multiple ensemble techniques employing diverse algorithms on the same datasets [29], these studies have typically not delved into the intricacies of interpretability. Consequently, a critical facet of AI model assessment—explicability—has been conspicuously absent from their work. This research endeavor distinguishes itself by addressing a notable gap in the current scholarly discourse. It contributes valuable insights into the vital statistics of explicability within the context of hybrid GAI-DML models. Through a rigorous examination of interpretability challenges and elucidation of their implications, this study enriches the existing academic literature and underscores the significance of explicability in facilitating the broader adoption of advanced AI systems within the business domain.

Various authors (e.g., [9], [23]) have presented comprehensive overviews of explainable and interpretable algorithms in the context of ML. They have highlighted the importance of model agnostic approaches to explainability, which can be used with a range of different MLs. Some authors have provided important insights into the use of hybrid generative-discriminative models for anomaly detection and their potential for improving explainability ([8], [28]). While some researchers [e.g., 28] have argued for the lack of solid explainability in such hybrid approaches, another group of studies [e.g., 8] demonstrated that the use of a hybrid approach can lead to improved performance compared to conventional models. This provides an argument to combine generative and discriminative models for anomaly detection systems, which requires further research for robust strategies, potentially increasing trust, and acceptance of AI in business applications. (e.g., [32]) have emphasized the importance of improving AI model explanations. They reviewed various aspects of XAI models and suggested new research directions. They also argued that interpretability should be an essential component of AI, requiring a deeper understanding of the underlying mechanisms and processes, beyond just providing predictions.

The development of XAI models, particularly for anomaly detection, is gaining interest. Traditional methods have improved performance but lack explainability. Further investigation into explainability is needed, particularly for hybrid models. A model-agnostic strategy can enhance the explainability of discriminative models.

## III. METHODOLOGICAL APPROACH

Deep learning has led to the rise of reconstruction methods for anomaly detection. These methods assume that a model trained on normal data will fail to reconstruct anomalous data, signaling the presence of anomalous data. Deep autoencoders (AE) have been used to develop reconstruction approaches for anomaly detection with remarkably superior results, but an expanding body of literature suggests even better outcomes when employing the more advanced and probabilistic variational autoencoders [19].

We propose the SVEAD framework, a multi-step approach to anomaly detection that starts by compressing data using VAE to a lower-dimensional space. Then, it leverages ensemble techniques to combine the outputs of various individual anomaly detection models. This process aims to provide a simplified yet comprehensive approach to identifying anomalies in complex datasets. Fig. 1 presents the proposed SVEAD framework.

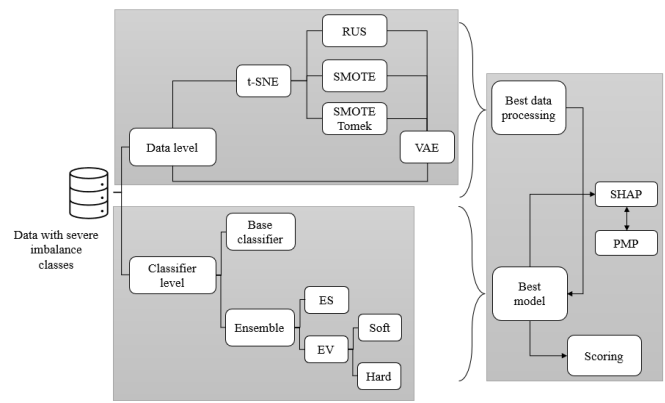

Fig 1. SVEAD Interpretable framework: Anomaly Detection (Source: Author)

The three-core approaches (t-SNE, VAE and EM) are discussed below to provide an argument and justification for using these methods.

### A. *t-distributed Stochastic Neighbor Embedding*

t-SNE is a dimensionality reduction technique that emphasizes the preservation of local data relationships. Its ability to uncover non-linear patterns and anomalies makes it a valuable tool in various fields, including anomaly detection. By projecting high-dimensional data into a lower-dimensional space, t-SNE helps to unmask hidden structures and deviations that might not be readily apparent in the original data space ([11]; [37]; [6]; [21]; [24]). It does so by employing a probabilistic approach to denote the similarity between data points and then minimizing the divergence (KL) between the high and low-dimensional similarity distributions.

$$p(i,j) = \left(\frac{p(i|j) + p(j|i)}{2N}\right) \quad (1)$$

$$q(i,j) = \frac{(1+||y_i - y_j||^2)^{-1}}{z} \quad (2)$$

$$KL\ divergence = \sum_i \sum_j p(i,j)\ log\frac{p(i,j)}{q(i,j)} \quad (3)$$

Equation (1) displays the working at high dimensional space, $p(i,j)$ = similarity between data points $i$ and $j$, $p(i|j)$ = conditional probability of choosing data point i as a neighbor of data point $j$, and $N$ = total number of data points. This equation calculates the similarity between data points andbased on conditional probabilities.

Equation (2) displays the working at low dimensional space, $q(i,j)$ = similarity between data points $i$ and $j$. It is computed based on Euclidean distance $||y_i - y_j||^2$ between data points in the lower-dimensional space, and it is normalized by $z$.

Equation (3) represents the KL divergence between the between the conditional probability distributions $p(i,j)$ and $q(i,j)$.

Fig. 2 displays a comparison of the clear separation between majority and minority classes by applying PCA, t-SVD and t-SNE to our dataset[1]. Researchers found that the t-SNE pipeline yields better visualization and is much better at preserving local structure ([19], [23]). So, we have experimented with t-SNE by extracting the embeddings generated by the algorithm. These are used as input features for downstream anomaly detection models.

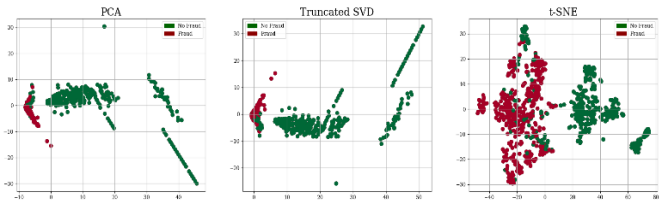

Fig. 2. Feature reduction (code concept taken from J M Bachmann blog)

t-SNE is non-deterministic and parameter-dependent, which is defined as perplexity. For a large sample size, the recommended perplexity value is between 20-30 ([26]).

### B. *Variational Auto Encoder*

The and of a distribution are represented by two vectors of size m that a VAE encoder learns to produce. From these vectors, a latent vector is sampled and transformed back to the original input vector.$m$ Several authors (e.g., [22]; [30]; [41]; [40]; [35]; [10]), have proposed the use of VAE for better interpretability and have demonstrated its effectiveness in different applications, including financial and healthcare data (e.g., [7] and [25]). This supports the argument that VAE can be a promising approach for anomaly detection. Moreover, it has been found that researchers [38] have applied VAE on the same dataset and obtained promising results which further strengthens this argument.

The encoder of the VAE maps the input data point x to a latent variable $z$, while the decoder maps $z$ to the reconstructed output $x'$. The encoder takes an input $x$ and computes the mean and std dev of the GD over $z$, presented in Equation (4):

$$q(z|x) = N(z;\ \mu(x), \sigma^2(x)I) \quad (4)$$

Equation (4) displays that $\mu(x)$ = mean, $\sigma^2(x)$ = std dev, represents the conditional probability distribution of the latent variable z given the input data x. In a VAE, this distribution is typically assumed to be GD (). The decoder takes z sampled from the GD and generates a reconstructed output x':

$$p(x'|z) = Bernoulli\ (x';f(z)) \quad (5)$$

Equation (5) displays that $f(z) = NN$, taking $z$ as input, and outputs the parameters of a BD over the reconstructed output $x'$. The objective function is to maximize the ELBO, which is defined in Equation (6):

$$ELBO = E\ [log\ p\ (x\ |\ z)] - KL\ [q\ (z\ |\ x)\ ||\ p(z)] \quad (6)$$

$E\ [log\ p(x\ |\ z)]$ = expected log-likelihood of the reconstructed output given the latent variable, and $KL\ [q\ (z\ |\ x)\ ||\ p(z)] = KL$ divergence between the encoder distribution and the prior distribution over the latent variable. The prior distribution is set to GD, $p(z) = N(z;\ 0, I)$. The ELBO can be re-written as in Equation (7) and subsequently in Equation (8):

$$ELBO = E\ [log\ p\ (x\ |\ z)] - KL\ [q\ (z\ |\ x)\ ||\ p(z)] \quad (7)$$

$$= E\ [log\ p\ (x\ |\ z)] - E\ [log\ q\ (z\ |\ x)] + E\ [log\ p(z)] - E\ [log\ p\ (z\ |\ x)] \quad (8)$$

The above equations consist of 4-terms:

- 1st term $(E\ [log\ p\ (x\ |z)]) \rightarrow$ reconstruction error,
- 2nd term $(KL[q\ (z\ |\ x)\ ||\ p\ (z)]) \rightarrow$ this is the divergence between the approximate posterior and prior distribution where, $q\ (z\ |\ x)$ = posterior and the $p(z)$ = prior distribution.
- 3rd term, $E\ [log\ p(z)] \rightarrow$ prior distribution of the latent variable $z$. The interaction between the prior distribution of the latent variable $z$ and the VAE's objective to bring the learned distribution in line with this prior distribution is the central for creating a meaningful and effective latent space. This ensures that the VAE learns structured and organized representations of data in the latent space, which in turn enables the model to perform various tasks, such as generating new data, manipulating existing data attributes, and detecting anomalies in the data.
- 4th term, $-E[log\,p(z|x)]$ → negative log-likelihood of the approximate posterior distribution q $(z|x)$ under the prior distribution $p(z)$. This term serves as a regularization component during training. It encourages the VAE to have its learned distribution (approximate posterior) of latent variable z be as close as possible to the specified prior distribution. In other words, it penalizes the model if it tries to represent data in a way that significantly deviates from the initially assumed distribution.

---

[1] The dataset was collected and analyzed as part of a research cooperation between Worldline and ULB's Machine Learning Group (http://mlg.ulb.ac.be) on big data mining and fraud detection.

To learn a useful latent representation of the input data, VAE tries to optimize the target function during training.

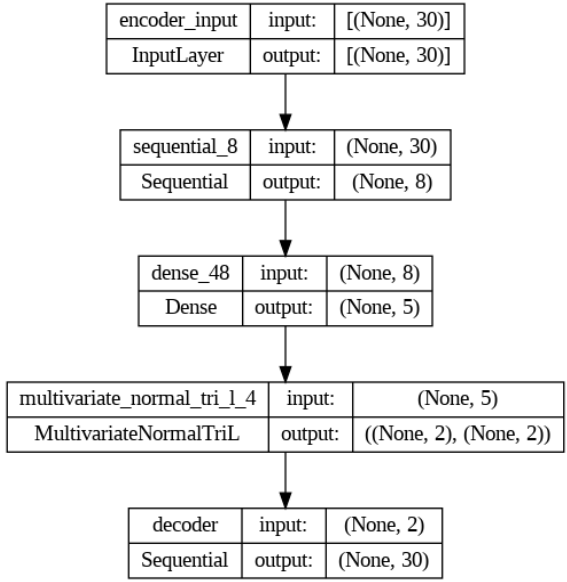

Fig. 3. Trained VAE

Reconstruction approaches for anomaly detection enhance confidence by identifying large reconstruction errors. Fig. 3 displays the network architecture that detects anomalous behavior, learning five distribution parameters for feature-independent normal distributions (two mean values and three covariance values).

MCS were conducted to estimate reconstruction probability, with 100 samples for each input. Fig. 4 shows distinct clustering across all t-SNE plots of latent distribution parameters and samples.

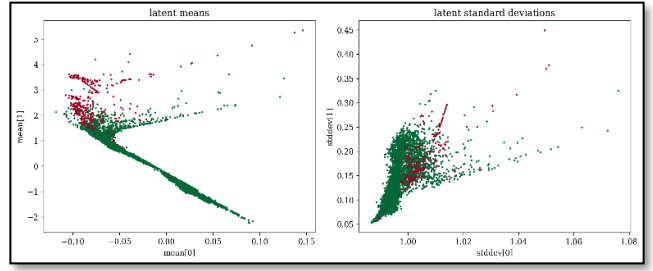

Fig. 4. t-SNE scatterplots: Latent Representations

The VAE is learning useful information from a distinct split between fraudulent and legitimate transactions, with fraudulent transactions being more dispersed and having larger values on both axes. This is consistent with the theory that anomalous transactions are mostly unpredictable.

*C. Ensemble Model*

The ensemble model implemented here combines various models to increase overall performance and mitigate the weaknesses of individual models. Assuming the target function $f(x_i, y_i)$ is the training data, the goal is to learn a hypothesis function $\hat{H}f(x)$ to approximate the target function $f$ as closely as possible. $\hat{H}f(x)$ is expressed as a weighted sum of the output of the base classifiers, as shown in Equation (9):

$$\hat{H}f(x) = w1 * h1(x) + w2 * h2(x) + \ldots + wn * hn(x) \quad (9)$$

where, $h_n(x)$ is the output of the $n^{th}$ base classifier, and $w_n$ is a weight determining the contribution of each base classifier to the final prediction. The weights $w_n$ learned by minimizing a loss function measure the difference between the predicted output of the ensemble and the true output. While EM can significantly enhance prediction accuracy, the combination of multiple models complicates the ability to explain the ensemble's decisions. Balancing model performance and interpretability is an ongoing challenge in the field of machine learning, especially when dealing with complex models like ensembles.

IV. DATA ANALYSIS & MODEL DEVELOPMENT

The information in the data includes fraudulent purchases made by cardholders in Europe. There were 492 fraudulent purchases (0.17%) and 28,4315 safe purchases (99.83%). It has 28 main components and is PCA encoded to ensure confidentiality. We have experimented with different supervised algorithms as the base learners, e.g., Log Regn, SVC, KNN, and RF. Table 2 displays the accuracy scores of all the CV models. CV ensures that the model does not overfit the training data and generalizes well to new data.

The last row of Table 2 displays the ROCAUC score, which identifies the best model to distinguish between fraud and safe transactions. Table 3 shows the learning curve analysis for a Log Reg model with varied training set sizes (30%, 60%, and 90% of the entire dataset). We can conclude from the findings that as the training set grows, the average training accuracy reduces significantly while the average test accuracy increases.

TABLE 2. LEARNING CURVE ANALYSIS

| Description | Log Reg | KNN | SVC | RF |
|---|---|---|---|---|
| Mean cv scores: original data | 0.9331 | 0.9183 | 0.9210 | 0.8929 |
| Mean cv scores: hyperparameters optimized | 0.9357 | 0.9237 | 0.9291 | 0.9197 |
| ROCAUC | 0.9751 | 0.9211 | 0.9751 | 0.9175 |

The high accuracy of the test dataset indicates that the model is operating well and generalizing to new data.

TABLE 3. LEARNING CURVE ANALYSIS

| Samples used to train model | Average train accuracy | Average test accuracy |
|---|---|---|
| 224 | 0.97 | 0.93 |
| 448 | 0.96 | 0.94 |
| 672 | 0.96 | 0.94 |

Fig. 5 displays the graphical representation of models' performance, showing overfitting or underfitting for selected hyperparameters.

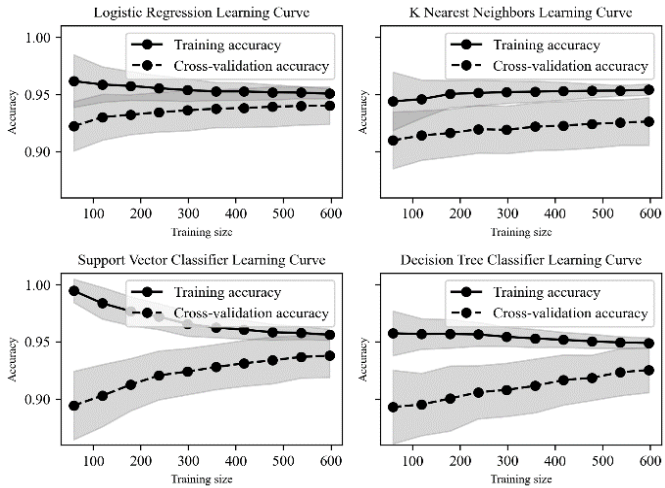

Fig. 5. Learning curves (code concept taken from J M Bachmann blog)

Three different sampling techniques were employed to compare the optimal output: Random Undersampler, SMOTE Oversampler, and combined sampler SMOTE with Tomek links. To avoid data leakage, it was crucial to sample after the CV. Fig. 6 displays the sampling pipeline employed for this work.

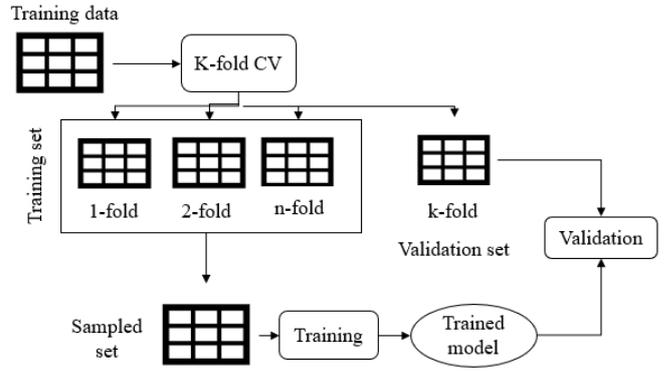

Fig. 6. Sampling approach during CV (Source: Author)

A. *Ensemble techniques*

We used simple ensemble techniques:

- Voting based: Hard Voting and Soft Voting.
- Stacking: Where all 4 models are combined in a hierarchical manner and finally their predictions are used as input to a meta-model (final_estimator), which produces the final prediction.

Fig. 7 displays the fitted ES architecture.

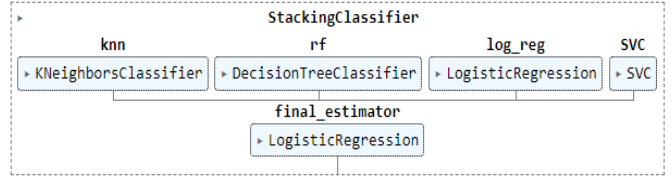

Fig. 7. Ensemble Stacking architecture.

The predictions from the basic estimators are combined in the final_estimator, which is a Log Reg model. It determines which basic models work well in scenarios based on training data patterns and modifies their weights. CV is used to avoid overfitting and using smaller base model sets can help mitigate overfitting. We also ensured that during meta-learning, estimator is trained on predictions that were not used during the training of base models to avoid data leakage.

B. *Evaluation criteria*

We used the following evaluation metrics for evaluating the ensemble models: Precision, Recall, F1-score, ROCAUC, PRAUC, MCC, CK, and BS.

To determine the best configuration for the provided dataset, the study investigated various VAE architectures. A sparse 2-layer overcomplete VAE with linear activation and dropout was found to perform well in dealing with the dataset's features, including noise. The VAE uses normal samples and standardized training data sets, creating a new EM training set for the estimator and enhancing performance by reducing data dimensionality and noise. Fig. 8 displays a scatter plot of the latent vectors obtained from encoding with the well separated classes in the latent space; we can see distinct clusters of points for each class. This supports the claim that the proposed framework can identify the distinct separate clusters of points for each class.

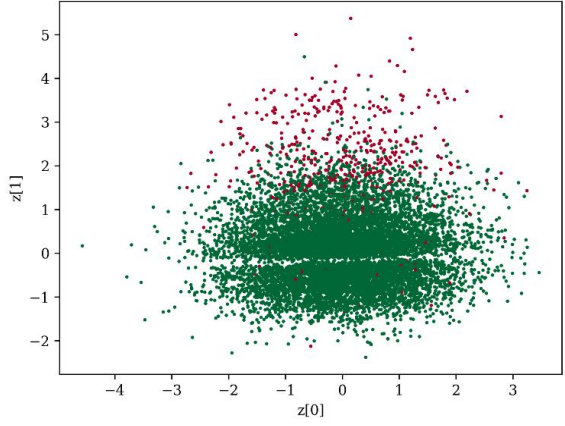

Fig. 8. Latent vector samples: Variational Auto Encoder

## V. SHAPLEY ADDITIVE EXPLANATIONS

SHAP provides a way to decompose the predictions of the model into individual contributions from each feature. It is model agnostic and helps identify key features driving the output and how they interact with each other. To determine an overall SHAP value for each feature in the ensemble, the SHAP values are computed for each individual model and the second level model and aggregated using a weighted average.

$$\varphi_i^S(x) = \sum_{T \subseteq s\{i\}} \frac{|T|!(|S|-|T|-1)!}{|S|!} * \{f_T(x_{T \cup \{i\}}) - f_T x_T\} \quad (10)$$

Where, $\varphi_i^S(x) =$ SHAP value of feature i, for instance, x, conditioned on a set of features S, $f_T(x_{T \cup \{i\}}) =$ model's output when features T and i are present in the input, $f_T x_T =$ model's output when only features T are present in the input and all other features are set to their background values, $|T| =$ cardinality of the set T (i.e., the number of features in T), and $|S| =$ cardinality of the set S (i.e., the number of features in S).

The summation in the equation goes over all subsets T of S that do not include feature i. The weighting factor inside the summation makes sure that the SHAP values satisfy several desired criteria, such additivity and consistency, and it depends on the size of T and S.

## VI. RESULTS & DISCUSSIONS

The objective of using VAE in architecture is to provide an interpretable latent space. By compressing the data into a lower-dimensional representation, VAE identifies the underlying structure of the data and, subsequently, the anomalous transactions.

Various accuracy measures were used by previous researchers working on imbalanced datasets, e.g., accuracy, recall, precision, TPR, FPR, specificity, and G-mean [12]; MCC, F2 Score, Kappa Score, Brier Score, and Precision [2]; accuracy, precision, recall, F1 score [44]; Precision, and AUPRC [26]. We opted for a combination of all those displayed in Table 4. All the models were trained and assessed using a 70%–30% train-test split, and various evaluation metrics were used to evaluate the models. The ES method has the best overall performance, displaying the highest precision score of 94.87% and the highest F1 score of 84.09%, suggesting a good balance between precision and recall. Additionally, the ROCAUC score of 87.75% indicates that it is the most effective at distinguishing between the two classes. ES is the meta-model of all the models trained on the same training dataset.

SMOTETomek + VAE + ES has the highest values for precision, recall, F1, ROCAUC, and AUPRC, indicating that it has the best overall performance among all models. It also has a high MCC and Kappa score, suggesting good agreement between predicted and actual labels. This model correctly identified 99.91% of positive cases. The precision score of 0.9878 indicates that the model is correctly identifying positive samples while minimizing the number of false positives. The MCC score of 0.898 indicates the model has the best overall performance in balancing precision, recall, and MCC.

SMOTE + VAE + Log Reg has the second-highest values for precision, recall, F1, ROCAUC, and AUPRC and has the highest MCC and Kappa score among all models. These metrics indicate that it has exceptionally good overall performance, although slightly lower than the top-performing model. For the SVEAD framework, the SHAP values are calculated for each model in the ensemble and the second-level model and combined using a weighted average to generate an overall SHAP value for each feature in the ensemble.

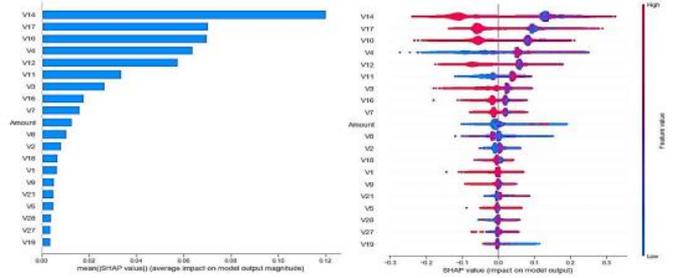

Fig. 9. Feature importance plot

Fig. 9 displays the feature importance plot of the variables by using PIP. V14, V17, and V10 are the top three features that influence model prediction and, thus, fraudulent, and non-fraudulent transactions. For a thorough knowledge of feature relevance, PIP is employed as a supplemental technique to SHAP. It shuffles the values of a single feature in the dataset at random before reevaluating the model's performance with the shuffled feature. By plotting the decrease in model performance against the features, a PIP plot is generated to rank the features based on their importance.

ICE plots can be used to create more localized explanations for a single individual [15]. Based on this argument, we further used the ICE plot to interpret the feature importance of the trained model.

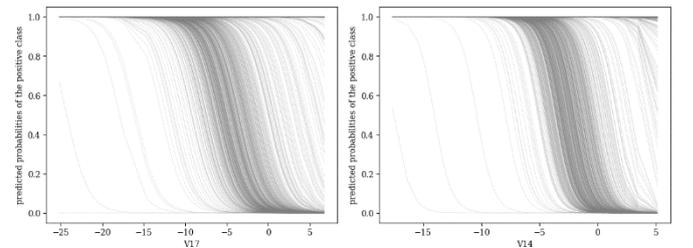

Fig 10. ICE plots of feature V17 & V14

Fig. 10 displays the ICE plots of features 'V14' and 'V17'. Each gray line represents the predicted probability of the positive class for a single observation, as the features are varied. As the feature value increases, the predicted probability increases, leading to the "waterfall" shape of the plot. The shape of each ICE curve reflects how the model's predicted probabilities for the positive class change as the features vary for each individual observation. The curves take a sigmoid-like shape, with the probability increasing or decreasing sharply at certain points, indicating that changes in the feature have a significant effect on the predicted probabilities. The sigmoid curve represents nonlinearities in the relationship between the features and the target variable.

The output of the model is not simply the result of a single algorithm, but a combination of the outputs of all the models. So, the question of explaianbility appears here, where we have implemented the model-agnostics SHAP framework into the architecture.

TABLE 4. ACCURACY METRICS

| Model[a] | Precision | Recall | F1 | ROCAUC | AUPRC | MCC | Kappa | BS |
|---|---|---|---|---|---|---|---|---|
| Log Reg | 0.8902 | 0.8801 | 0.8725 | 0.8723 | 0.8523 | 0.7981 | 0.7995 | 0.0982 |
| RUS + Log Reg | 0.8947 | 0.9015 | 0.8981 | 0.9636 | 0.9692 | 0.8155 | 0.8321 | 0.0718 |
| ROS + Log Reg | 0.9434 | 0.9126 | 0.9420 | 0.988 | 0.990 | 0.8894 | 0.8877 | 0.0402 |
| t-SNE + Log Reg (SMOTE) | 0.9113 | 0.8921 | 0.8183 | 0.9036 | 0.9099 | 0.8179 | 0.8178 | 0.0579 |
| VAE + Log Reg + RUS | 0.9224 | 0.9149 | 0.9186 | 0.9736 | 0.9799 | 0.8379 | 0.8378 | 0.0579 |
| SMOTE + VAE + Log Reg | 0.9495 | 0.9361 | 0.9583 | 0.9720 | 0.9721 | 0.9193 | 0.9183 | 0.0299 |
| t-SNE+ EV (SMOTE) | 0.8933 | 0.6836 | 0.7745 | 0.8417 | 0.7888 | 0.7811 | 0.7742 | 0.1272 |
| t-SNE + ES (RUS) | 0.9302 | 0.8163 | 0.8695 | 0.9081 | 0.9111 | 0.8279 | 0.8118 | 0.1351 |
| RUS + VAE +ES | 0.9778 | 0.9677 | 0.9835 | 0.9121 | 0.9002 | 0.8345 | 0.8762 | 0.1392 |
| SMOTE + VAE + ES | 0.9878 | 0.9871 | 0.9824 | 0.9789 | 0.9741 | 0.8280 | 0.8258 | 0.1552 |
| SMOTETomek + VAE + ES | 0.9898 | 0.9891 | 0.9894 | 0.9892 | 0.9799 | 0.8980 | 0.8354 | 0.1675 |

Statistics quantifies the reliability, validity, and uncertainty of AI models and offers understandable justifications for their predictions. However, mathematically optimal statistical approaches may not always be computationally feasible, and efficient data analysis techniques may not always be statistically optimal [42]. The argument also mentions that there is growing regulatory pressure to ensure that ML models are transparent and understandable as their use spreads. This highlights the importance of XAI; using XAI techniques, we can enhance the interpretability of our model, which can help in ensuring the responsible and ethical use of AI model.

VII. LIMITATIONS & FUTURE DIRECTIONS

User-centered design approaches for XAI systems employed in anomaly detection would be the subject of future research. This would entail researching the integration of user feedback and domain knowledge into the model-building and explanation-generation processes. Moreover, the concerns of long-term model maintenance, such as those relating to concept drift, data quality, and model retraining, are not addressed in the study. Practical methods for sustaining XAI-enhanced anomaly detection systems throughout time would be the subject of future research. Moreover, the study focuses primarily on the development and assessment of a hybrid anomaly detection framework, with a significant emphasis on technical issues and performance measures. However, there has been little discussion of the actual issues, constraints, and considerations that need to be addressed when deploying such a framework in real-world scenarios. A future direction would be to fill these research gaps.

CONCLUSIONS

The SVEAD framework is a powerful tool for anomaly detection, achieving high performance while maintaining interpretability and explainability. It uses diverse base models and ensemble stacking to optimize collective performance. The pre-processing stage includes SMOTETomek and VAE to balance imbalanced classes and extract relevant features. The VAE component provides a low-dimensional representation of input data, making it easier to understand. The ensemble model is analyzed individually to identify key features for anomaly detection. However, for large datasets, the complexity, ensemble methods, and SHapley Additive exPlanations values may result in processing overhead and scalability concerns.